\documentclass[runningheads]{llncs}
\usepackage{graphicx}
\usepackage{comment}
\usepackage{amsmath,amssymb} 
\usepackage{color}

\usepackage{algorithm, algorithmic}
\usepackage{amsfonts}
\usepackage{dsfont}
\usepackage{color}
\usepackage{bm}
\usepackage{multirow}

\newcommand{\Wb}{\mathbf{W}}

\newcommand{\xb}{\mathbf{x}}

\newcommand{\hb}{\mathbf{h}}

\newcommand{\Nc}{\mathcal{N}}

\begin{document}
\pagestyle{headings}
\mainmatter
\def\ECCVSubNumber{3386}  

\title{Learning Graph-Convolutional Representations for Point Cloud Denoising} 


\titlerunning{Learning Graph-Convolutional Representations for Point Cloud Denoising}
%
\author{Francesca Pistilli\inst{1}\orcidID{0000-1111-2222-3333} \and
Giulia Fracastoro\inst{1}\orcidID{0000-0001-8495-1097} \and
Diego Valsesia\inst{1}\orcidID{0000-0003-1997-2910} \and
Enrico Magli\inst{1}\orcidID{0000-0002-0901-0251}}
\authorrunning{F. Pistilli et al.}
%
\institute{Politecnico di Torino, Italy\\
\email{francesca.pistilli@polito.it}
}
\maketitle

\begin{abstract}
Point clouds are an increasingly relevant data type but they are often corrupted by noise. We propose a deep neural network based on graph-convolutional layers that can elegantly deal with the permutation-invariance problem encountered by learning-based point cloud processing methods. The network is fully-convolutional and can build complex hierarchies of features by dynamically constructing neighborhood graphs from similarity among the high-dimensional feature representations of the points. When coupled with a loss promoting proximity to the ideal surface, the proposed approach significantly outperforms state-of-the-art methods on a variety of metrics. In particular, it is able to improve in terms of Chamfer measure and of quality of the surface normals that can be estimated from the denoised data. We also show that it is especially robust both at high noise levels and in presence of structured noise such as the one encountered in real LiDAR scans.
\end{abstract}

\section{Introduction}

A point cloud is a geometric data type consisting in an unordered collection of 3D points representing samples of 2D surfaces of physical objects or entire scenes. Point clouds are becoming increasingly popular due to the availability of instruments such as LiDARs and the interest in exploiting the richness of the geometric representation in challenging applications such as autonomous driving. However, the acquisition process is imperfect and a significant amount of noise typically affects the raw point clouds. Therefore, point cloud denoising methods are of paramount importance to improve the performance of various downstream tasks such as shape matching, surface reconstruction, object segmentation and more. 

Traditional model-based techniques \cite{alexa2003computing,oztireli2009feature,guennebaud2007algebraic,lipman2007parameterization,huang2013edge,cazals2005estimating} have typically focused on fitting a surface to the noisy data. Such techniques work well in low-noise settings but they usually suffer from oversmoothing, especially in presence of high amounts of noise or geometries with sharp edges. Given the success of learning-based methods, in particular those exploiting deep neural networks, in a wide variety of tasks, including image denoising and restoration problems \cite{zhang2017beyond,liu2018non}, a few works have recently started exploring point cloud denoising with deep neural networks. The most challenging problems in processing point clouds are the lack of a regular domain, such as a grid, and the fact that a point cloud is just a set of points and any permutation of them still represents the same data. Any learning-based method must therefore learn a \textit{permutation-invariant} function that can deal with data defined on an irregular domain. This is a significant challenge that point cloud processing algorithms tackled by either approximating the irregular domain with a grid, e.g. by building voxels, or building a permutation-invariant function as a composition of operations acting on single points (e.g., size-1 convolution) and a globally symmetric function (e.g., a max pool) as done by PointNet \cite{qi2017pointnet}. Neither of these solutions is completely satisfactory. The former introduces an undesirable approximation, while the latter lacks the expressiveness of convolutional neural networks (CNN) where the convolution operation extracts features that are localized as functions of the neighborhood of a pixel and features of features are assembled in a hierarchical manner by means of multiple layers, progressively expanding the receptive field. Recently, graph convolution \cite{bronstein2017geometric} has emerged as an elegant way to build operators that are intrinsically permutation-invariant and defined on irregular domains, while also exploiting some of the useful properties of traditional convolution, such as localization and compositionality of the features as well as efficient weight reuse. In particular, spatial-domain definitions of graph convolution have been recently applied in several problems involving point clouds such as classification \cite{simonovsky2017dynamic}, segmentation \cite{wang2019dynamic}, shape completion \cite{litany2018deformable} and generation \cite{valsesia2018learning}. Notably, the point cloud denoising problem has yet to be addressed with graph-convolutional neural networks.

In this paper, we propose a deep graph-convolutional neural network for denoising of point cloud geometry. The proposed architecture has an elegant fully-convolutional behavior that, by design, can build hierarchies of local or non-local features to effectively regularize the denoising problem. This is in contrast with other methods in the literature that typically work on fixed-size patches or apply global operations \cite{rakotosaona2019pointcleannet,duan20193d}. Moreover, dynamic computation of the graph from similarities among the high-dimensional feature-space representations of the points allows to uncover more complex latent correlations than defining neighborhoods in the noisy 3D space. Extensive experimental results show a significant improvement over state-of-the-art methods, especially in the challenging conditions of high noise levels. The proposed approach is also robust to structured noise distributions such as the ones encountered in real LiDAR acquisitions.

\section{Related work}\label{rel_work}
The literature on 3D point cloud denoising is vast and it can be subdivided into four categories: local surface fitting methods \cite{alexa2003computing,oztireli2009feature,guennebaud2007algebraic,lipman2007parameterization,huang2013edge,cazals2005estimating}, sparsity-based methods \cite{avron2010l,sun2015denoising,mattei2017point}, graph-based methods \cite{zeng20183d,dinesh20183d,schoenenberger2015graph}, and learning-based methods \cite{rakotosaona2019pointcleannet,hermosilla2019total,duan20193d,roveri2018pointpronets}. Among the methods belonging to the first category, the moving least squares (MLS) approach \cite{alexa2003computing} and its robust extensions \cite{oztireli2009feature,guennebaud2007algebraic} are the most widely used. Other surface fitting methods have also been proposed for point cloud denoising, such as jet fitting \cite{cazals2005estimating} or parameterization-free local projector operator (LOP) \cite{lipman2007parameterization,huang2013edge}. These methods achieve remarkable performance at low levels of noise, but they suffer from over-smoothing when the noise level is high \cite{han2017review}.

A second class of point cloud denoising methods \cite{avron2010l,sun2015denoising,mattei2017point} is based on sparse representations. In this case, the denoising procedure is composed of two minimization problems with sparsity constraints, where the first one estimates the surface normals and then the second one uses them in order to update the point positions. However, at high levels of noise the normal estimation can be very poor, leading to over-smoothing or over-sharpening \cite{sun2015denoising}.

Another approach for point cloud denoising is derived from the theory of graph signal processing \cite{shuman2013emerging}. These methods \cite{zeng20183d,dinesh20183d,schoenenberger2015graph} first define a graph whose nodes are the points of the point cloud. Then, graph total variation (GTV)-based regularization methods are applied for denoising. These techniques have proved to achieve very strong performance when the noise level is low. Instead, at high noise levels, the graph construction can become unstable, negatively affecting the denoising performance.

In the last years, learning-based methods \cite{rakotosaona2019pointcleannet,hermosilla2019total,duan20193d,roveri2018pointpronets}, especially the ones based on deep learning, have been gaining attention. Extending convolutional neural networks to point cloud data is not straightforward, due to the irregular positioning of the points in the space. However, in the context of shape classification and segmentation, many methods have recently been proposed specifically to handle point cloud data. PointNet \cite{qi2017pointnet} is one of the most relevant works in this field, where each point is processed independently before applying a global aggregation. Recently, a few methods proposed to extend the approach of PointNet to point cloud denoising. PointCleanNet \cite{rakotosaona2019pointcleannet} uses an approach similar to PointNet in order to estimate correction vectors for the points in the noisy point cloud. Instead, in \cite{duan20193d} the authors use a neural network similar to PointNet to estimate a reference plane for each noisy point and then they obtain the denoised point cloud by projecting the noisy point onto the corresponding reference plane. Also PointProNet \cite{roveri2018pointpronets} performs point cloud denoising by employing an architecture similar to PointNet in order to estimate the local directions of the surface. However, the main drawback of  these techniques based on PointNet is that they work on individual points and then apply a global symmetric aggregation function, but they do not exploit the local structure of the neighborhood. PointCleanNet addresses this issue by taking as input local patches instead of the entire point cloud. However, this solution is still limited by the fact that the network cannot learn hierarchical feature representations, like standard CNNs.

Graph-convolutional networks have shown promising performance on tasks such as segmentation and classification. In particular, DGCNN \cite{wang2019dynamic} first introduced the idea of a dynamic graph update in the hidden layers of a graph-convolutional network. However, the denoising problem is significantly different from the classification and segmentation tasks addressed in \cite{wang2019dynamic}, that rely more on global features instead of localized representations. In particular, there are several design choices that make DGCNN unsuitable for point cloud denoising: the spatial transformer block is not useful for denoising since it seeks a canonical global representation, whereas denoising is mostly concerned with local representations of point neighborhoods and also significantly increases the computational complexity for large point clouds; the graph convolution operation uses a max operator in the aggregation, which is unstable in presence of noise; the specific graph convolution definition is also less general than the one presented in this paper, which allows to implement adaptive filters where the aggregation weights are dependent on the feature vectors instead of being fixed as in \cite{wang2019dynamic}, as well as incorporating an edge attention term which is especially important in presence of noise because it promotes a lowpass behavior by penalizing edges with large feature variations.

\section{Proposed method}
In this section we present the proposed Graph-convolutional Point Denoising Network (GPDNet), i.e., a deep neural network architecture to denoise the geometry of point clouds based on graph-convolutional layers. The focus of the paper is to investigate the potential of graph convolution as a simple and elegant way of dealing with the permutation invariance problem encountered when processing point clouds. For this reason, we focus on analyzing the network in a discriminative learning setting where a clean reference is available and it is perturbed with white Gaussian noise. We refer the reader to \cite{hermosilla2019total} for a technique to train any point cloud denoising network in an unsupervised fashion only using noisy data.

\subsection{Architecture}

\begin{figure*}
    \centering
    \includegraphics[width=0.9\textwidth]{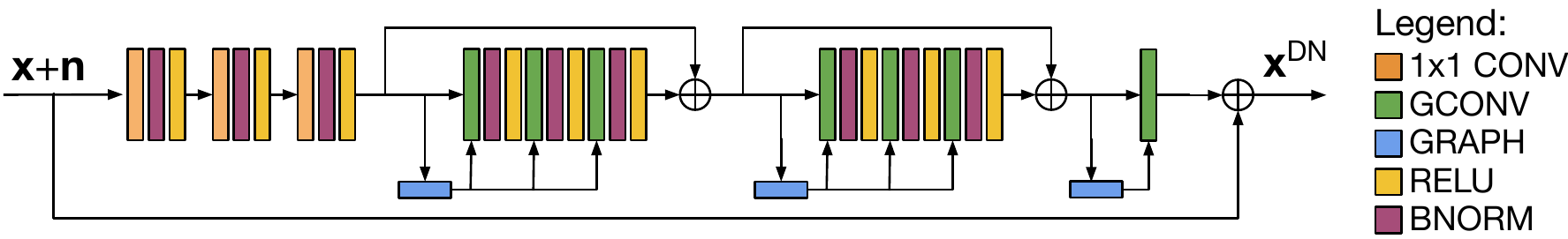}
    \caption{GPDNet: graph-convolutional point cloud denoising network.}
    \label{fig:net}
\end{figure*}

An overview of the architecture of GPDNet is shown in Fig. \ref{fig:net}. At a high level, it is a residual network that estimates the noise component of the input point cloud, which has been shown \cite{zhang2017beyond} to be easier than directly cleaning the data. A first block is composed of three single-point convolutions that gradually transform the 3D space into an $F$-dimensional feature space. Then a cascade of two residual blocks is used, having an input-output skip connection to reduce vanishing gradient issues. Each residual block is composed of three graph-convolutional layers. The graph is computed by selecting the $k$ nearest neighbors to each point in terms of Euclidean distances in the feature space. Notice that the graph construction is dynamic, i.e., it is updated after every residual block but shared among the graph convolutional layers inside the block to limit computational complexity. Dynamic construction of a similarity graph has been shown to induce more powerful feature representations \cite{wang2019dynamic,valsesia2018learning} and, in the context of a residual denoising network, it progressively uncovers the latent correlations that have not yet been eliminated. Intuitively, dynamic graph construction is preferable over building the graph in the noisy 3D space as neighborhoods might be strongly perturbed at high noise variances, leading to unstable or sub-optimal neighbor assignments.
All layers are interleaved with batch normalization which stabilizes training, especially in presence of Gaussian noise.
Finally, the last graph-convolutional layer projects the features back to the 3D space. 

\subsection{Graph-convolutional layer}

The core of the proposed architecture is the graph-convolutional layer. Graph convolution is a generalization of convolution to data that are defined over the nodes of a general graph rather than a grid. Multiple definitions of graph convolution have been proposed to capture salient properties of classical convolution, notably localization and weight reuse. In this paper, we use a modified version of the Edge-Conditioned Convolution (ECC) \cite{simonovsky2017dynamic} to address vanishing gradient and over-parameterization. In particular, we use some of the approximations introduced in \cite{valsesia2019deep} in the context of image denoising. 

The graph-convolutional layer has two inputs: a tensor representing a feature vector for each point, and a graph where nodes are points and edges represent similarities between points. The output feature vector $\hb_i^{l+1} \in \mathbb{R}^{F^l}$ of point $i$ at layer $l$ is computed by performing a weighted aggregation over its neighborhood $\Nc_i^l$ as defined by the graph:
\begin{align}
\hb_i^{l+1} = \Wb^l \hb_i^l + \sum_{j\in\Nc_i^l} \gamma^{l,j\to i} \frac{\sum_{t=1}^r\omega_t^{j\to i}\bm{\phi}_t^{j\to i} \bm{\psi}_t^{j\to i^T}\hb_j^l}{|\Nc_i^l|}.
\end{align}
The weights include a self-loop matrix $\Wb^l \in \mathbb{R}^{F^{l+1}\times F^l}$ which is shared among all points. The other weights in the aggregation, i.e., vectors $\bm{\phi}_t^{j\to i} \in \mathbb{R}^{F^{l+1}}$, $\bm{\psi}_t^{j\to i} \in \mathbb{R}^{F^{l}}$ and scalar $\omega_t^{j\to i}$ are computed as functions of the difference between the feature vector of point $i$ and point $j$, i.e., $\bm{\phi}_t^{j\to i}, \bm{\psi}_t^{j\to i}, \omega_t^{j\to i} = \mathcal{F}\left( \hb_i^l - \hb_j^l \right)$. This function is implemented as a multilayer perceptron (MLP) with two layers, where the final fully-connected layer can be approximated by means of a stack of circulant matrices since the number of free parameters would otherwise be very large. The value $r$ is a hyperparameter setting the maximum rank of the aggregation weight matrix obtained by explicitly computing $\sum_{t=1}^r\omega_t^{j\to i}\bm{\phi}_t^{j\to i} \bm{\psi}_t^{j\to i^T}$, again to reduce the number of parameters and memory requirements of the aggregation operation.  
The parameter $\gamma^{l,j\to i}$ is a scalar edge attention term which exponentially depends on the Euclidean distance between feature vectors across an edge:
\begin{align}
    \gamma^{l,j\to i} = \exp\left( - \Vert \hb_i^l - \hb_j^l \Vert^2_2 / \delta \right),
\end{align}
being $\delta$ a decay hyperparameter.

This definition of graph convolution has some advantages over alternative definitions such as GraphSAGE \cite{hamilton2017inductive}, FeastNet \cite{verma2018feastnet} or DGCNN \cite{wang2019dynamic}. In particular, the aggregation weights are functions of feature differences making the filtering operation performed by the graph-convolutional layer \textit{adaptive}. Moreover, since the function is implemented as an MLP, it can be more general than a fixed function with some learnable parameters.

The graph is constructed by searching for the $k$-nearest neighbors of each point in terms of Euclidean distance between their feature vectors. To limit complexity, a search area of predefined size, centered around the point, is defined, e.g., as a fixed number of neighbors in the noisy 3D space (see Fig. \ref{fig:recfield} for a visual representation of the search area and feature space neighborhoods).

We remark that GPDNet is fully-convolutional thanks to the graph convolution operation. By fully-convolutional we mean that the output feature vector of each point at a given layer is obtained as a multi-point aggregation of the feature vectors of neighboring points in the previous layer, thus building complex hierarchies of aggregations. This is in contrast with PointCleanNet \cite{rakotosaona2019pointcleannet} which works by processing each patch independently to estimate the denoised version of the central point. 
That approach does not create hierarchies of features obtained by successive multi-point aggregations, as in a classical CNN. The graph-convolutional structure recovers this behavior and can learn more powerful feature spaces.

\subsection{Loss functions}\label{loss_fcn}
We consider two loss functions to train the proposed method in a supervised setting. The first one is the mean squared error (MSE) between the denoised point cloud $\hat{\xb}$ and its noiseless ground truth $\xb$, i.e.:
\begin{align}
    L_\mathrm{MSE} = \frac{1}{N} \sum_{i=1}^N \Vert \hat{\xb}_i - \xb_i \Vert_2^2
\end{align}
being $N$ the number of points in the point cloud. This is the most natural choice in presence of Gaussian noise. However, it does not exploit prior knowledge about the distribution of points. In fact, it does not use the fact that the points may lie on a surface and therefore the tangential component of the noise is not as relevant as the normal component.

This property can be incorporated by regularizing the MSE loss with a term measuring the distance of the denoised point from the ground truth surface. Such measure can be approximated by the proximity to surface metric which computes the distance between the denoised point and the closest ground truth point. The loss function (MSE-SP) then becomes:
\begin{align}
    L_\mathrm{MSE-SP} = \frac{1}{N} \sum_{i=1}^N \left[ \Vert \hat{\xb}_i - \xb_i \Vert_2^2 + \lambda \min_{j} \Vert \hat{\xb}_i - \xb_j \Vert_2^2  \right]
\end{align}
for a regularization hyperparameter $\lambda$. Other works also considered proximity to surface in the loss function. Notably, PointCleanNet \cite{rakotosaona2019pointcleannet} uses a loss that combines the proximity to surface with a dual term measuring the distance between a ground truth point and the closest denoised point. This is done to ensure that the denoised points do not collapse into filament structures. We found that using the MSE to enforce this property provides better results.

\section{Experimental results}
In this section an experimental evaluation against state-of-the-art approaches as well as an analysis of the proposed technique is performed. Code is available online\footnote{https://github.com/diegovalsesia/GPDNet}.

\subsection{Experimental setting}
The training and test set are created selecting post-processed subsets of ShapeNet \cite{shapenet} repository.
This database is composed by 3D models of 55 object categories, each one described by a collection of meshes. Before utilization, the data has to be sampled and normalized.
First we sample 30720 uniformly distributed points for each model, then we rescale the obtained point clouds normalizing their diameter in order to ensure that data are at the same scale.
More than 100000 patches of 1024 points each are randomly selected from the point clouds to create the training set, taking point clouds from all the categories except 10 reserved for the test set.
Each patch is created by randomly selecting a point from a point cloud and collecting its 1023 closest points. 
The test set is constituted by 100 point clouds taken from ten different categories: airplane, bench, car, chair, lamp, pillow, rifle, sofa, speaker, table. We randomly select ten models from each category and sample 30720 uniformly distributed points from each model.\par
GPDNet is trained for a fixed noise variance for approximately 700000 iterations, each one characterized by a batch size of 16. The number of features used for all the layers is 99, except for the first three single-point convolutional layers where the number of features is gradually increased from 33 to 66 and finally to 99. The Adam optimizer has been employed with a fixed learning rate equal to $10^{-4}$. Concerning the graph-convolutional implementation, the rank $r$ for the low-rank approximation is set to 11, 3 circulant rows are considered for the construction of the circulant matrix, and $\delta=10$. 
During testing, GPDNet takes as input the whole point cloud and a search area is associated to each point of the point cloud, wherein the neighbors are searched and identified. Unless otherwise stated, 16 nearest neighbors in terms of Euclidean distances are used for graph construction.

\subsection{Comparisons with state-of-the-art}



\begin{table*}[]
\caption{Chamfer measure ($\times 10^{-6}$), $\sigma=0.01$, 16-NN.}
\setlength\tabcolsep{2.5pt} 
\vspace{-0.2cm}
\scriptsize{
\begin{tabular}{ccccccccccc}
Class    & Noisy    &DGCNN& APSS   & RIMLS  & AWLOP    & MRPCA    & GLR      & PCN & \textbf{GPDNet} & \textbf{GPDNet} \\
    &     & \cite{wang2019dynamic}   & \cite{guennebaud2007algebraic}  &   \cite{oztireli2009feature}  &   \cite{huang2013edge}  &\cite{mattei2017point}  &\cite{zeng20183d} & \cite{rakotosaona2019pointcleannet}& \textbf{MSE} & \textbf{MSE-SP} \\\hline \hline
airplane & 50.32 & 44.82 & 28.22 & 39.73 & 31.27 & 28.19 & 19.56 & 26.36 & \textbf{17.22} & 17.58  \\ \hline
bench    & 48.71 &38.70 & 26.97 & 32.76 & 34.08 & 32.93 & 20.43 & 27.64 & \textbf{19.33} & 19.80  \\ \hline
car      & 64.34 & 60.47 & 47.73 & 55.56 & 54.21 & 44.33 & 42.22 & 75.34 & \textbf{38.09} & 38.14  \\ \hline
chair    & 60.78 & 59.69 &37.31 & 45.65 & 47.91 & 38.41 & 34.98 & 55.10 & \textbf{29.50} & 29.69  \\ \hline
lamp     & 59.73 & 52.54 &24.57 & 34.02 & 35.23 & 31.51 & 19.67 & 20.58 & \textbf{16.17} & 17.15  \\ \hline
pillow   & 69.79 & 64.28&\textbf{15.64} & 21.23 & 46.36 & 23.95 & 17.59 & 21.07 & 17.11 & 19.04  \\ \hline
rifle    & 38.97 & 26.99 &36.01 & 49.37 & 27.79 & 23.49 & 15.84 & 15.09 & 14.45 & \textbf{14.00}  \\ \hline
sofa     & 69.63 & 65.05 & \textbf{22.27} & 28.04 & 53.08 & 32.14 & 30.88 & 43.36 & 25.87 & 27.21  \\ \hline
speaker  & 73.50 & 68.72 &\textbf{26.50} & 30.19 & 58.92 & 47.57 & 40.78 & 76.09 & 34.87 & 35.81  \\ \hline
table    & 56.21 & 50.17 &27.45 & 32.63 & 41.26 & 34.78 & 27.12 & 43.02 & \textbf{24.27} & 24.64  \\ \hline
\end{tabular}}
\label{C2C_0.01}
\end{table*}


\begin{table*}[]
\caption{Chamfer measure ($\times 10^{-6}$), $\sigma=0.015$, 16-NN.}
\setlength\tabcolsep{2.5pt} 
\vspace{-0.2cm}
\scriptsize{
\begin{tabular}{ccccccccccc}
Class    & Noisy    & DGCNN & APSS   & RIMLS  & AWLOP    & MRPCA    & GLR      & PCN & \textbf{GPDNet} & \textbf{GPDNet} \\
    &     & \cite{wang2019dynamic}   & \cite{guennebaud2007algebraic}  &   \cite{oztireli2009feature}  &   \cite{huang2013edge}&   \cite{mattei2017point}  &\cite{zeng20183d} & \cite{rakotosaona2019pointcleannet}& \textbf{MSE} & \textbf{MSE-SP} \\\hline \hline
airplane & 97.78 & 84.40 & 86.42 & 106.33 & 73.32 & 67.39 & 36.76 & 35.27 & 28.47 & \textbf{27.62}  \\ \hline
bench    & 94.82 & 64.76 & 75.51 & 91.93 & 82.04 & 70.05 & 32.19 & 30.10 & 28.72 & \textbf{26.96}   \\ \hline
car      & 102.23 & 93.43 & 72.56 & 103.52 & 93.38 & 69.88 & 55.92 & 92.23 & 52.92 & \textbf{51.77} \\ \hline
chair    & 105.16 & 94.4 5& 81.47 & 104.38 & 92.47 & 73.45 & 48.62 & 69.18 & 46.28 & \textbf{43.73} \\ \hline
lamp     & 120.65 & 112.06 & 65.79 & 82.40 & 88.78 & 77.09 & 39.93 & 30.59 & \textbf{27.37} & 28.60  \\ \hline
pillow   & 132.57 & 113.32 & 22.74 & 42.54 & 112.54 & 73.67 & 31.38 & 29.02 & \textbf{23.32} & 27.25 \\ \hline
rifle    & 80.40 & 61.04 & 92.14 & 110.51 & 69.35 & 55.65 & 31.81 & \textbf{21.45} & 28.43 & 22.48  \\ \hline
sofa     & 121.02 & 99.63 & 42.80 & 69.92 & 107.58 & 72.62 & 51.12 & 61.15 & \textbf{40.10} & 42.04 \\ \hline
speaker  & 123.27 & 114.12 & 46.45 & 58.28 & 110.29 & 77.95 & 53.75 & 87.68 & \textbf{49.20} & 49.57 \\ \hline
table    & 103.50 & 84.95 & 62.64 & 78.21 & 89.33 & 70.87 & 37.94 & 43.88 & 36.06 & \textbf{33.89}  \\ \hline
\end{tabular}}
\label{C2C_0.015}
\end{table*}


\begin{table*}[]
\caption{Chamfer measure ($\times 10^{-6}$), $\sigma=0.02$, 16-NN.}
\setlength\tabcolsep{2.5pt} 
\vspace{-0.2cm}
\scriptsize{
\begin{tabular}{ccccccccccc}
Class    & Noisy    & DGCNN& APSS   & RIMLS  & AWLOP    & MRPCA    & GLR      & PCN & \textbf{GPDNet} & \textbf{GPDNet} \\
    &     & \cite{wang2019dynamic}   & \cite{guennebaud2007algebraic}  &   \cite{oztireli2009feature}  &   \cite{huang2013edge}  &\cite{mattei2017point} & \cite{zeng20183d} & \cite{rakotosaona2019pointcleannet}& \textbf{MSE} & \textbf{MSE-SP} \\\hline \hline
airplane & 161.79 & 127.44 & 175.68 & 186.24 & 145.94 & 123.71 & 90.55 & 74.17 & 45.96 & \textbf{42.30}  \\ \hline
bench    & 161.52 & 99.36 & 166.85 & 182.42 & 157.29 & 127.51 & 83.99 & 90.34 & 41.24 & \textbf{36.77}  \\ \hline
car      & 148.74 &113.94 & 141.69 & 167.78 & 145.51 & 109.49 & 77.56 & 160.08 & 72.06 & \textbf{67.43} \\ \hline
chair    & 163.75 & 132.91 & 160.01 & 155.38 & 158.12 & 122.70 & 79.85 & 145.56 & 67.91 & \textbf{60.16} \\ \hline
lamp     & 204.05 & 153.02 & 178.08 & 198.22 & 187.31 & 146.41 & 109.24 & 85.31 & 45.21 & \textbf{44.60} \\ \hline
pillow   & 215.58 & 190.32 & 164.83 & 196.53 & 206.14 & 150.65 & 85.86 & 92.84 & \textbf{34.47} & 38.58  \\ \hline
rifle    & 144.18 & 131.91 & 195.68 & 176.07 & 144.22 & 105.87 & 89.19 & 71.57 & 43.07 & \textbf{29.55}  \\ \hline
sofa     & 184.11 & 155.51 & 166.34 & 190.91 & 178.93 & 133.98 & 89.31 & 144.72 & \textbf{62.58} & 65.06 \\ \hline
speaker  & 186.01 & 136.72 & 138.80 & 162.34 & 180.45 & 126.17 & 84.37 & 160.26 & 66.57 & \textbf{63.40} \\ \hline
table    & 168.32 & 115.00 &171.25 & 179.81 & 162.36 & 125.72 & 78.06 & 102.17 & 50.47 & \textbf{44.80} \\ \hline
\end{tabular}}
\vspace{-0.1cm}
\label{C2C_0.02}
\end{table*}

In this section the proposed method is compared with state-of-the-art methods for point cloud denoising. As described in Section \ref{rel_work}, different categories of point cloud denoising methods are present in the literature. In the experiments, we take into account at least one algorithm from each category. APSS \cite{guennebaud2007algebraic} and RIMLS \cite{oztireli2009feature} are well-known MLS-based surface fitting methods and they were tested using the MeshLab software \cite{LocalChapterEvents:ItalChap:ItalianChapConf2008:129-136}. AWLOP \cite{huang2013edge} is another surface fitting method and it was implemented using the software released by the authors. MRPCA \cite{mattei2017point} is a sparsity-based method and it was implemented using the code provided by the authors. GLR \cite{zeng20183d} is one of the most promising works belonging to the graph-based category and it was implemented using the code provided by the authors. PointCleanNet (PCN) \cite{rakotosaona2019pointcleannet} is one of the most recent learning-based methods and its code is publicly available. In order to ensure a fair comparison, PointCleanNet was retrained with additive Gaussian noise at a specific standard deviation, instead of using the blind model released by the authors. We also include a modified version of DGCNN \cite{wang2019dynamic} as an additional baseline. This modified version replaces the segmentation head with a single-point convolution to regress the point displacement. 

As metric to evaluate the performance of the proposed method, we compute the Chamfer measure, also called Cloud-to-Cloud (C2C) distance. This metric is widely utilized in point cloud denoising, because it computes an average distance of the denoised points from the original surface. First, the mean distance between each denoised point and its closest ground truth point is computed, then the one between each ground truth point and its closest denoised point. The Chamfer measure is then their average:
\begin{align}
    \mathrm{C2C} = \frac{1}{2N} \Bigg[ \sum_{i=1}^N \min_j \Vert \hat{\xb}_i - \xb_j \Vert_2^2 
    + \sum_{j=1}^N \min_i \Vert \hat{\xb}_i - \xb_j \Vert_2^2 \Bigg].
\end{align}
The results of the experiments at different noise levels are reported in Table \ref{C2C_0.01}, \ref{C2C_0.015} and \ref{C2C_0.02}. As described in Sec. \ref{loss_fcn}, in the proposed network we consider two different loss functions obtaining two versions of the proposed method, namely GPDNet MSE and GPDNet MSE-SP.
It is clearly visible that both versions of the proposed method significantly outperform state-of-the-art methods, especially at medium and high levels of noise, as shown in Table \ref{C2C_0.015} with $\sigma=0.015$ and Table \ref{C2C_0.02} with $\sigma=0.02$. Instead, at low noise level the other algorithms become more competitive and the performance gap decreases, but the proposed method still obtains the best results in the majority of the categories, as reported in Table \ref{C2C_0.01}. 
This can be explained by the fact that most of the other methods involves surface reconstruction or normal estimation, operations that cannot be computed with sufficient accuracy at high levels of noise. Instead, the proposed method directly estimates the denoised point cloud.
In addition, it can be observed from Table \ref{C2C_0.02} that the GPDNet MSE-SP version is particularly effective at high levels of noise, outperforming GPDNet MSE in almost all the categories. This behavior can be explained by the regularizing effect of the surface distance component of the loss, which is especially useful at high noise variance due to the fact that it can incorporate more prior knowledge about the data. The performance difference between the two variants decreases at low noise levels, as shown in Table \ref{C2C_0.01}. It is worth noting that DGCNN shows poor performance for the reasons explained in Sec. \ref{rel_work}, being originally designed for classification or segmentation. This is in line with the results presented in the PointCleanNet paper \cite{rakotosaona2019pointcleannet} where the authors also show the poor denoising performance of DGCNN, and highlights the importance of the design in this paper, which is tailored to the denosing task.

We also consider another metric for a quantitative assessment of the denoised point clouds.
In particular, we assess whether an off-the-shelf algorithm for surface normal estimation can produce more accurate normals when provided with point clouds denoised by the proposed method. Since surface normals are widely used in many applications, we believe that measuring their quality when extracted from the denoised data is a relevant metric for the characterization of a denoiser.
In this experiment we consider a different test set, composed of 5 well-known point clouds: Armadillo, Bunny, Column, Galera and Tortuga. The change of dataset is motivated by the availability of ground truth normals for these point clouds.
For every denoising method considered in the comparison, we compute the unoriented normal vector of each point in the denoised point cloud. The standard algorithm employed for the normal estimation is the built-in MATLAB function, which is based on principal component analysis.
We compute the unoriented normal angle error (UNAE) as
\vspace{-0.1cm}
\begin{align}
\mathrm{UNAE} = \frac{1}{N}\sum_{i=1}^N\mathrm{arcos} \Bigg[ 1-\frac{1}{2} \min &\Bigg( \|\hat{\mathbf{n}}_{i^*} - \mathbf{n}_i\|_2^2 ,\|\hat{\mathbf{n}}_{i^*} + \mathbf{n}_i\|_2^2\Bigg)\Bigg],
\vspace{-0.1cm}
\end{align}
where $\mathbf{n}_i$ is the groud-truth normal vector at $\xb_i$ and $\hat{\mathbf{n}}_{i^*}$ is the estimated normal vector at the denoised point closest to $\xb_i$.
Table \ref{un_norm} reports the average error across the five test point clouds.
A minimum error of about six degrees is reported since the MATLAB algorithm introduces a non-zero estimation error in the computation of the normals, as can be seen from the first column of Table \ref{un_norm}. It can be observed that the proposed denoising method, in particular the version with only MSE as loss function, increases the accuracy of the normal estimation, outperforming the state-of-the-art at each noise level considered. It is also interesting to notice that learning-based methods are more stable to noise than model-based methods as their performance degrades more gracefully for increasing noise variance.

\begin{table*}[t]
\caption{Unoriented normal angle error (degrees), 16-NN.}
\setlength\tabcolsep{2pt} 
\vspace{-0.2cm}
\scriptsize{
\begin{tabular}{cccccccccccc}
$\sigma$ & Clean & Noisy &DGCNN& APSS  & RIMLS & AWLOP & MRPCA & GLR   & PCN & \textbf{GPDNet} & \textbf{GPDNet} \\     
&  &   & \cite{wang2019dynamic}   &\cite{guennebaud2007algebraic}   & \cite{oztireli2009feature}  &   \cite{huang2013edge}  &   \cite{mattei2017point}  &\cite{zeng20183d} & \cite{rakotosaona2019pointcleannet}& \textbf{MSE} & \textbf{MSE-SP} \\\hline \hline
0.01     & 6.44      & 31.13 &30.83 &22.60 & 24.52 & 29.79 & 31.40 & 21.90 & 26.85         & \textbf{20.11}     & 22.33              \\ \hline
0.015    & 6.44      & 32.77 & 32.52 & 31.83 & 37.35 & 32.17 & 39.97 & 25.99 & 27.54         & \textbf{21.16}     & 24.46              \\ \hline
0.02     & 6.44      & 33.77 & 32.31& 42.42 & 45.86 & 33.41 & 42.45 & 31.30 & 28.65         & \textbf{22.78}     & 27.06              \\ \hline
\end{tabular}}
\vspace{-0.3cm}
\label{un_norm}
\end{table*}

\begin{figure*}[t]
\centering
\includegraphics[width=0.19\textwidth]{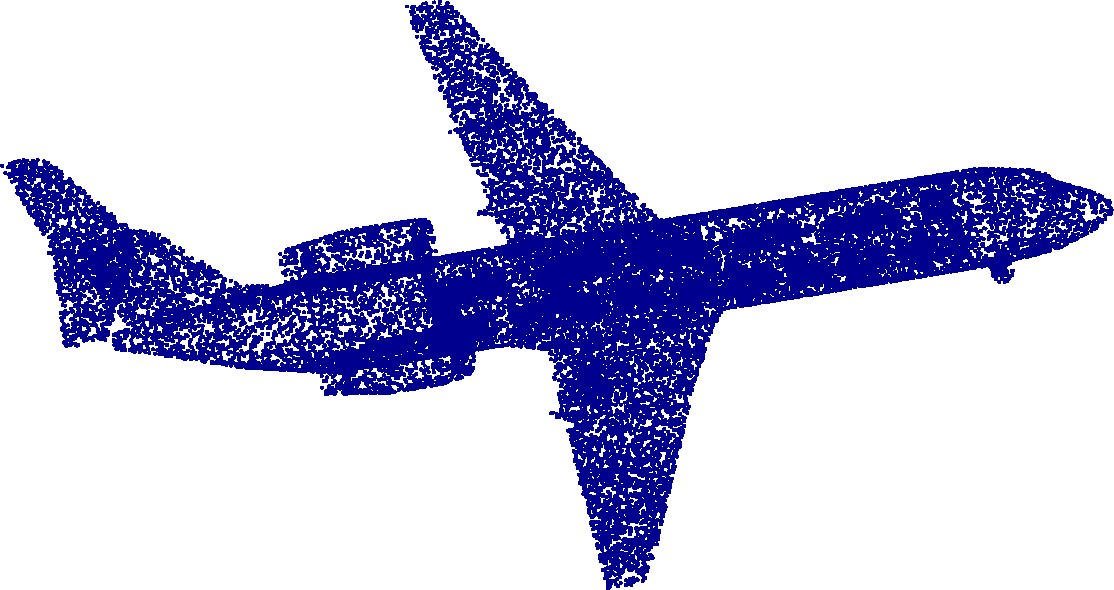}
\includegraphics[width=0.19\textwidth]{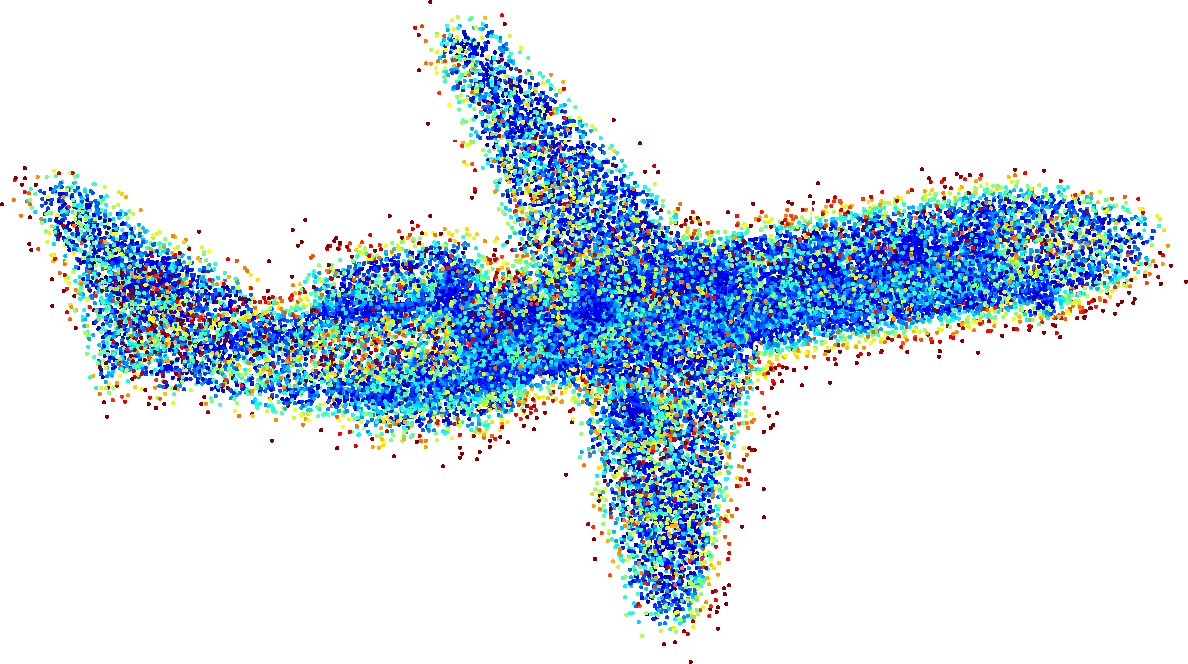}
\includegraphics[width=0.19\textwidth]{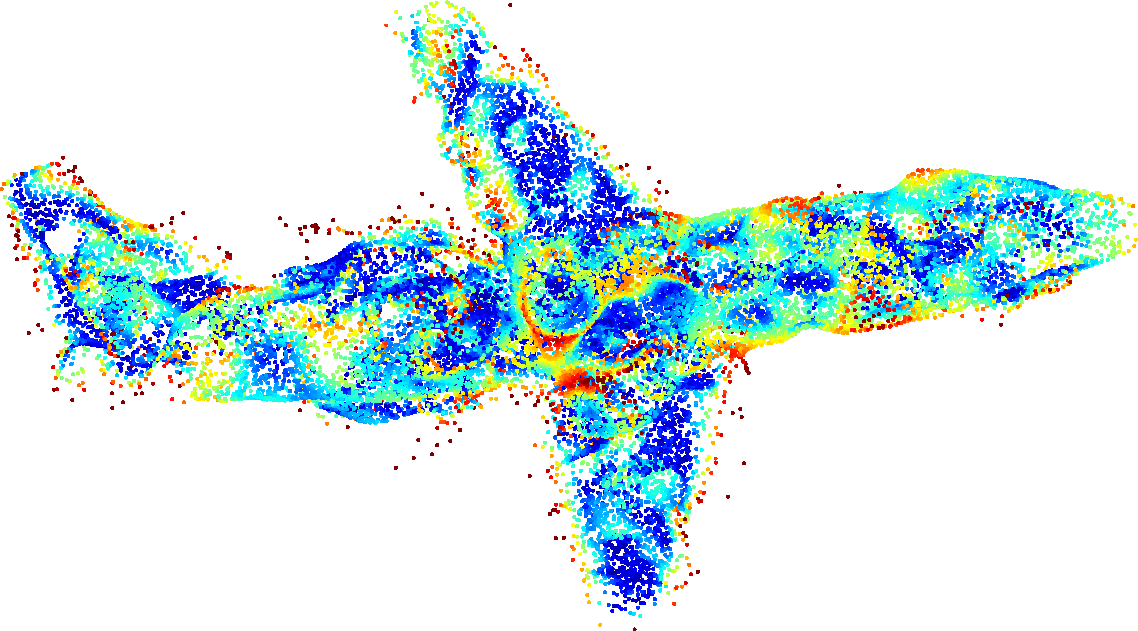}
\includegraphics[width=0.19\textwidth]{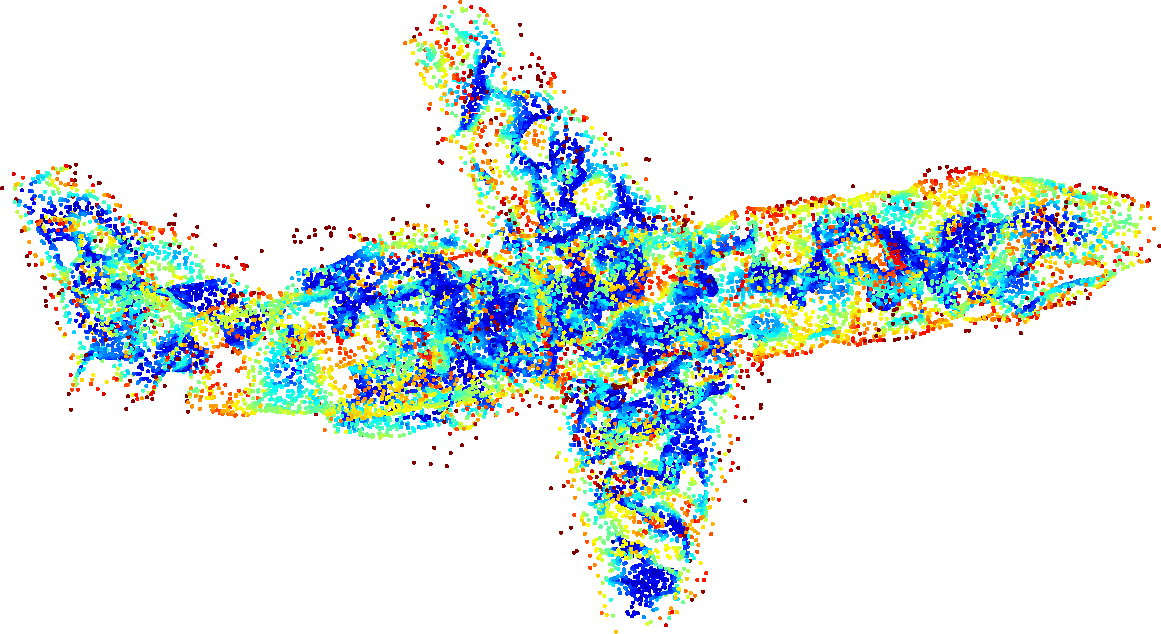}
\includegraphics[width=0.19\textwidth]{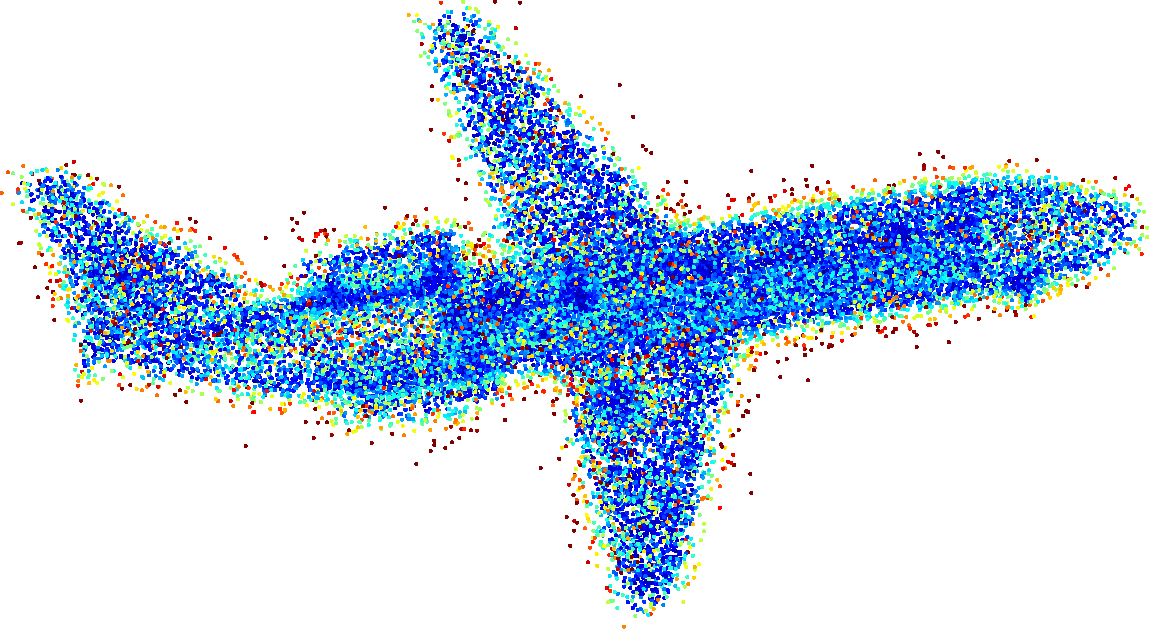}\\
\includegraphics[width=0.19\textwidth]{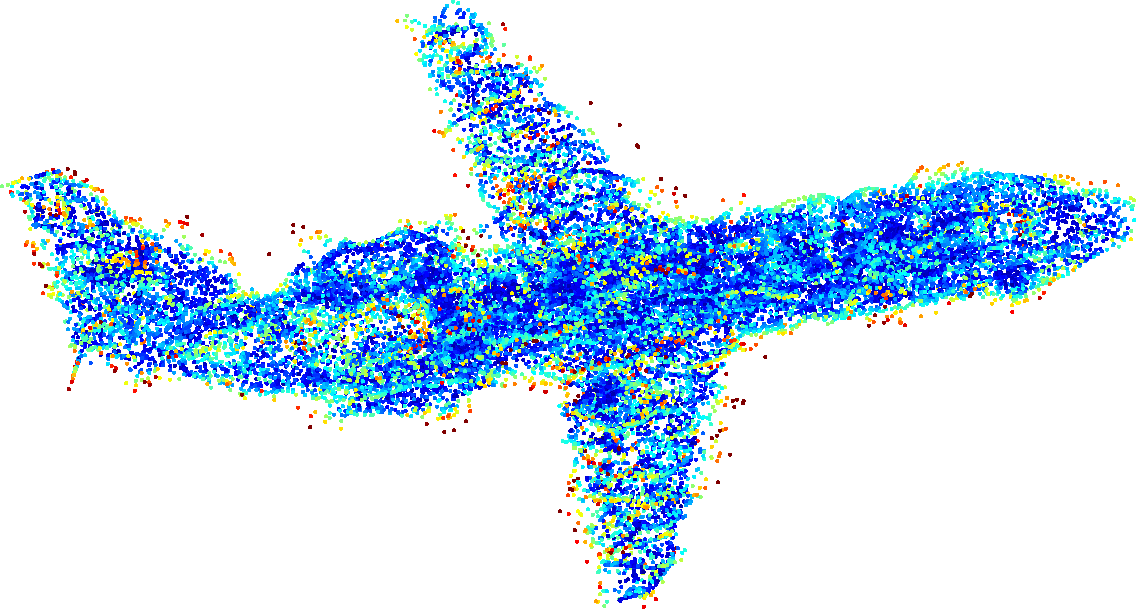}
\includegraphics[width=0.19\textwidth]{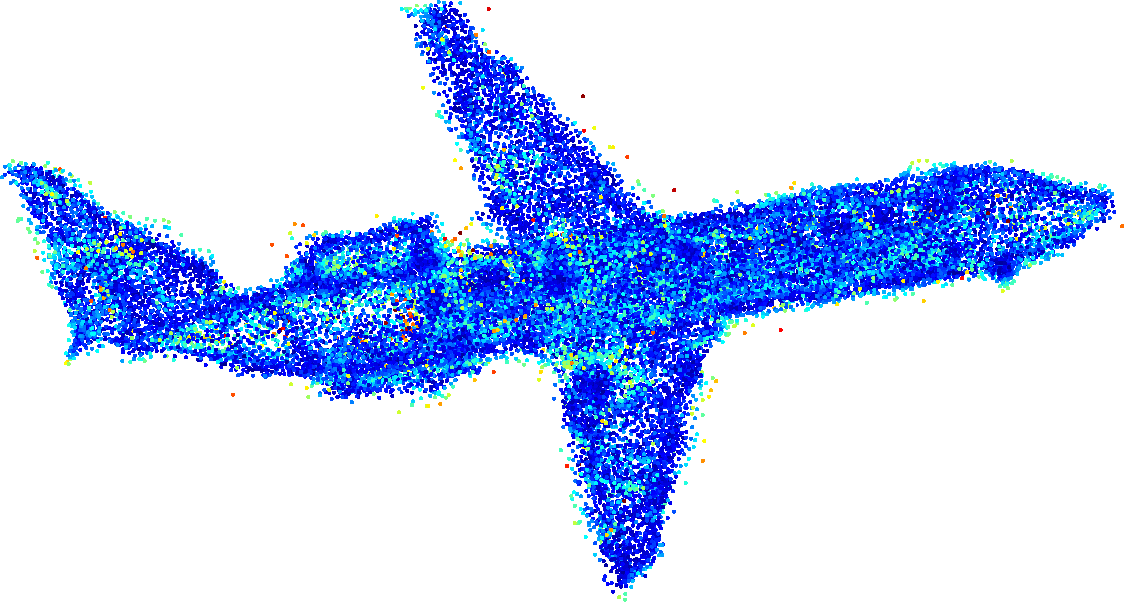}
\includegraphics[width=0.19\textwidth]{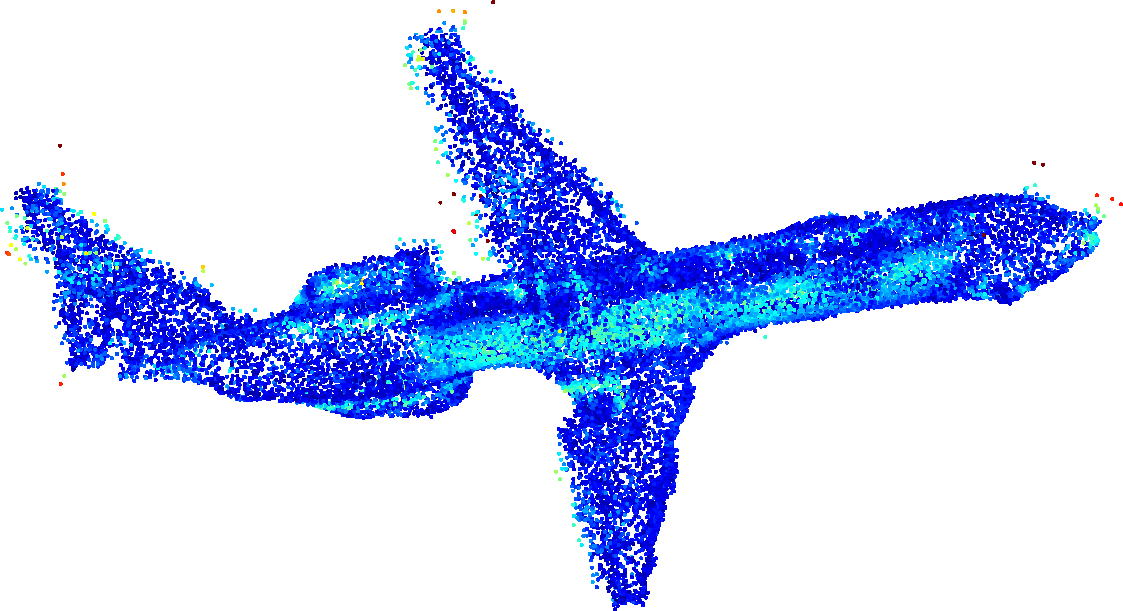}
\includegraphics[width=0.19\textwidth]{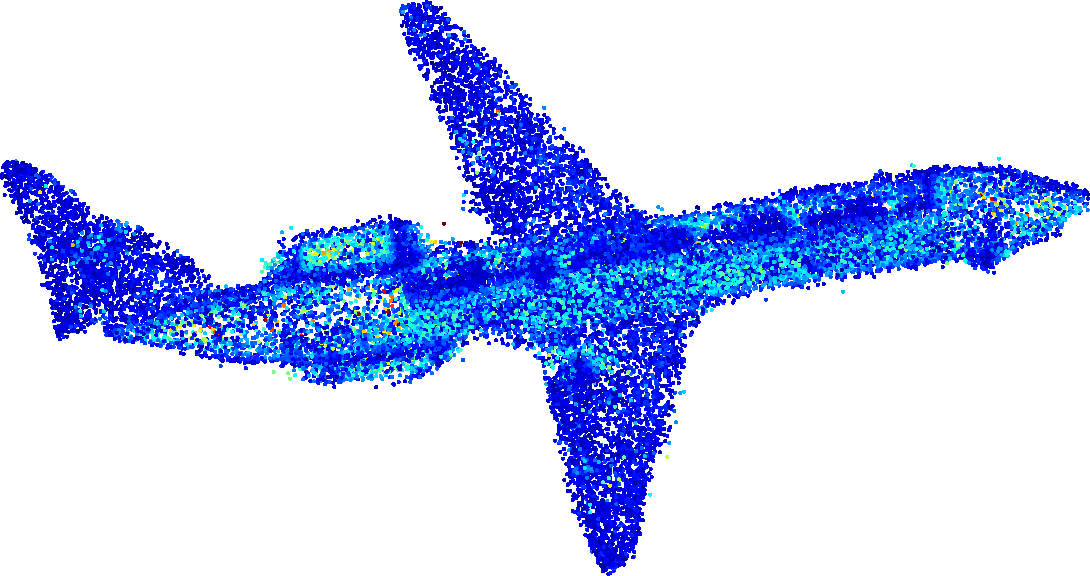}
\includegraphics[width=0.19\textwidth]{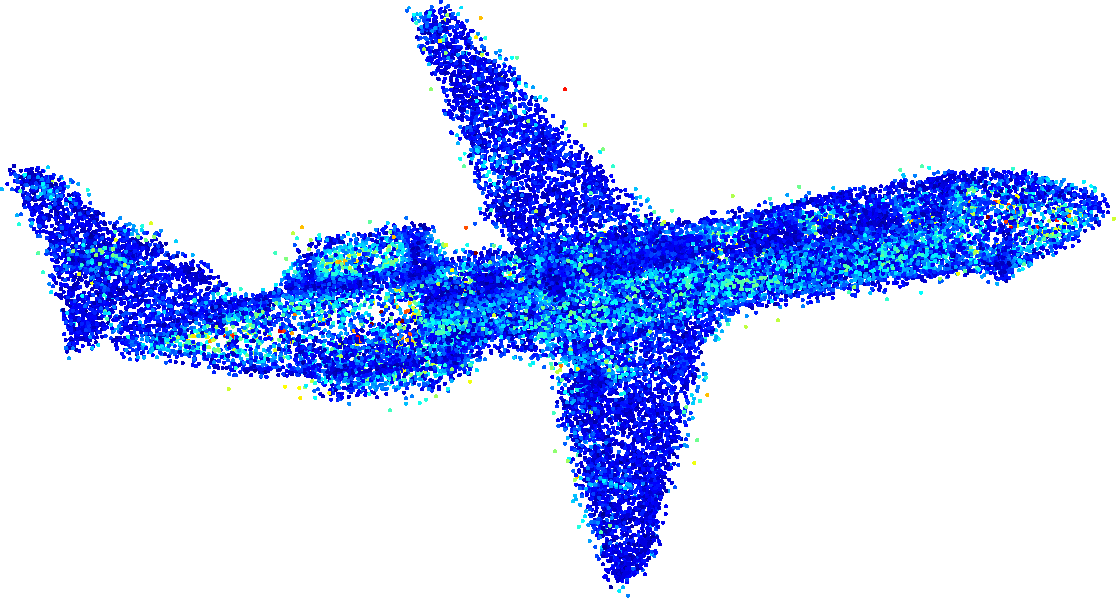}
\vspace{-0.12cm}
\caption{Denosing results for $\sigma=0.015$. Color represents distance to surface (red is high, blue is low). Top left to bottom right: clean point cloud, DGCNN ($\mathrm{RMSD}=0.0091$), APSS ($0.0123$), RIMLS ($0.0127$), AWLOP ($0.0106$), MRPCA ($0.0096$), GLR ($0.0070$), PointCleanNet ($0.0065$), GPDNet MSE ($0.0060$), GPDNet MSE-SP ($0.0062$).}
\vspace{-0.05cm}
\label{fig:denoised}
\end{figure*}

\begin{figure*}[t]
\centering
\includegraphics[width=0.19\textwidth]{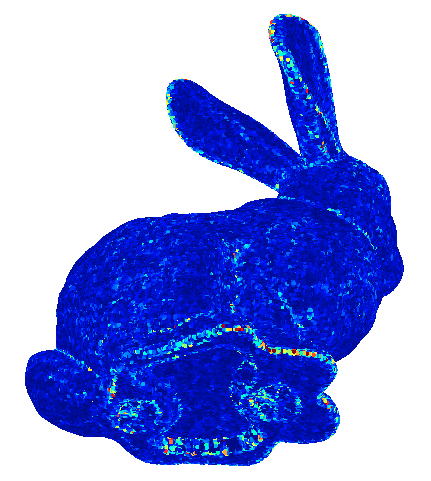}
\includegraphics[width=0.19\textwidth]{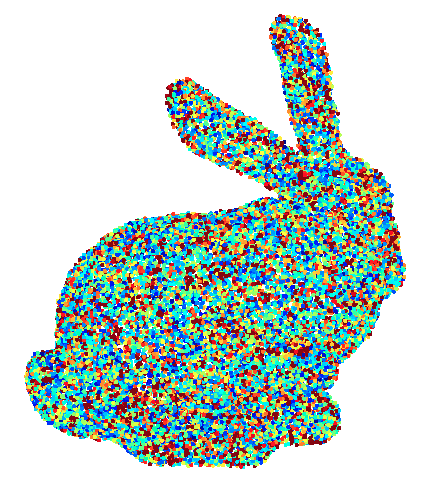}
\includegraphics[width=0.19\textwidth]{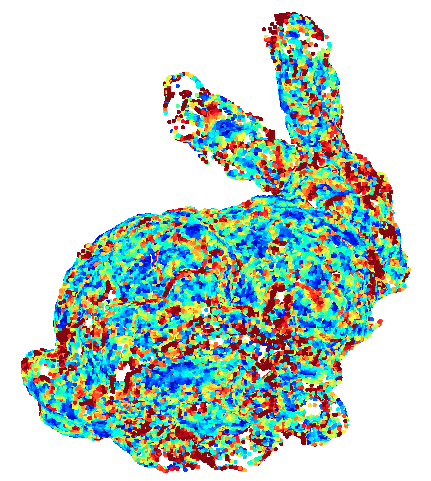}
\includegraphics[width=0.19\textwidth]{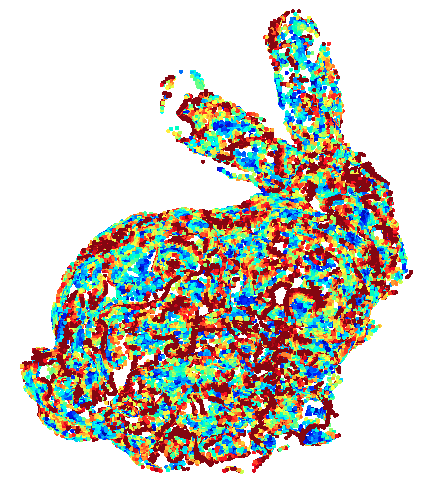}
\includegraphics[width=0.19\textwidth]{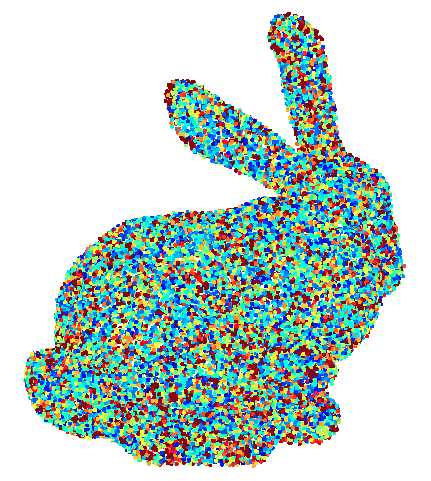}\\\vspace{-0.1cm}
\includegraphics[width=0.19\textwidth]{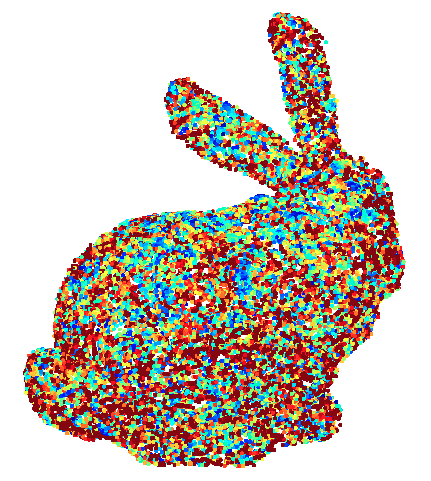}
\includegraphics[width=0.19\textwidth]{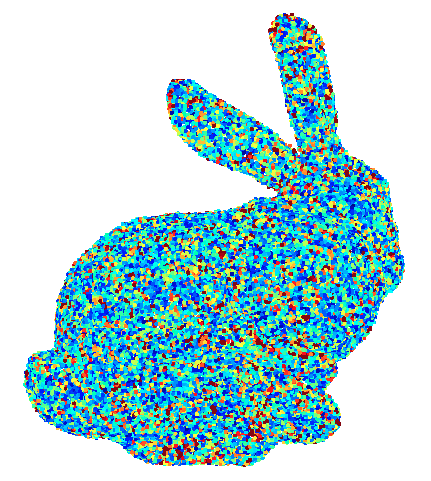}
\includegraphics[width=0.19\textwidth]{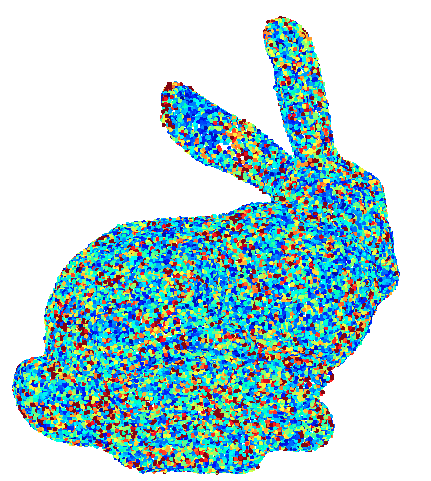}
\includegraphics[width=0.19\textwidth]{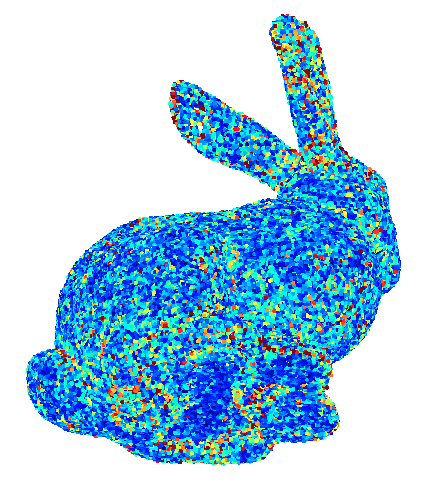}
\includegraphics[width=0.19\textwidth]{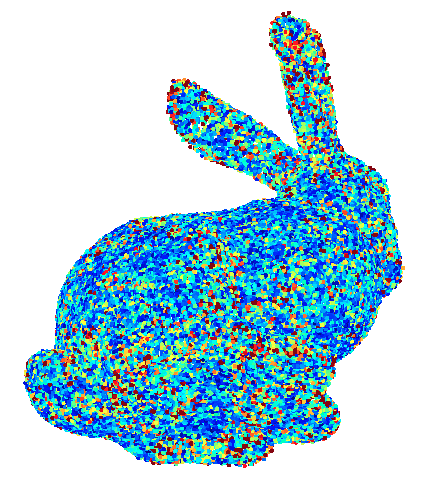}
\caption{Denosing results  for $\sigma=0.015$. Color represents unoriented normal angle error (red is high, blue is low). Top left to bottom right: clean point cloud ($\mathrm{UNAE}=3.75^\circ$), DGCNN ($29.73^\circ$), APSS ($26.29^\circ$), RIMLS ($33.63^\circ$), AWLOP ($29.18^\circ$), MRPCA ($37.50^\circ$), GLR ($22.08^\circ$), PointCleanNet ($23.63^\circ$), GPDNet MSE ($16.62^\circ$), GPDNet MSE-SP ($23.11^\circ$).}
\label{fig:denoised_norm}
\end{figure*}

Fig. \ref{fig:denoised} shows qualitative results at a medium noise level by presenting the denoised point cloud for each method. The surface distance of each point is visualized in the figure to understand the position of the denoised points with respect to the ground truth. The root mean square value of the surface distance (RMSD) can be computed as:
\begin{align}
    \mathrm{RMSD} = \sqrt{\frac{1}{N} \sum_{i=1}^N \min_j \Vert \hat{\xb}_i - \xb_j \Vert^2_2}.
\end{align}
It can be seen that on average both versions of our method provide lower points-surface distance and that the shape of the reconstructed point cloud is more similar to the original one. Fig. \ref{fig:denoised_norm} shows another qualitative comparison, displaying the unoriented normal estimation error for each denoised point. It can be seen that the proposed method, especially the version with only MSE, provides lower normal estimation errors, highlighting the higher quality of the denoised point cloud.

\subsection{Ablation studies}

We study the behavior of GPDNet in terms of a few design choices. In particular, we first investigate the impact of dynamic graph computation, i.e., updating the graph from the hidden feature space as in Fig. \ref{fig:net}, as opposed to a fixed graph construction where neighbors are identified in the noisy 3D space and used for all graph-convolutional layers. Table \ref{table:fixed_dynamic} shows that dynamic graph update provides improved performance thanks to refined neighbor selection.

We also study the impact of neighborhood size on the overall performance. Selecting a larger number of neighbors for graph convolution increases the size of the receptive field and can help denoise smooth areas in the point cloud by capturing more context, at the price of losing some localization and increased computational complexity. This is related to results on image denoising \cite{burger2012image}, where it is known that the optimal size of the receptive field depends on the noise variance. Tables \ref{table:nn} and \ref{table:nn_norm} show that increasing the number of neighbors is beneficial, up to a saturation point. We also see that the impact of a larger receptive field is more significant for the GPDNet MSE-SP variant.

\begin{table}[]
\vspace{-0.5cm}
\centering
\caption{Fixed vs. Dynamic graph, $\sigma=0.015$, 8-NN.}
\setlength\tabcolsep{5pt} 
\vspace{-0.1cm}
\begin{tabular}{lcccc}
                & \multicolumn{2}{c}{\textbf{GPDNet MSE}} & \multicolumn{2}{c}{\textbf{GPDNet MSE-SP}} \\ 
                & Dynamic                & Fixed         & Dynamic            & Fixed             \\ \hline \hline
C2C ($\times 10^{-6}$) & \textbf{35.68}         & 37.00         & \textbf{36.99}     & 38.45             \\ \hline
UNAE (degrees)    & \textbf{23.56}         & 23.75         & \textbf{26.29}              & 26.65    \\ \hline
\end{tabular}
\label{table:fixed_dynamic}
\vspace{-1cm}
\end{table}

\begin{table}[]
\centering
\caption{Number of neighbors (Chamfer measure $\times 10^{-6}$).}
\setlength\tabcolsep{5pt} 
\vspace{-0.1cm}
\begin{tabular}{clcccc}
  &  & 4-NN & 8-NN & 16-NN & 24-NN \\ \hline \hline
\multirow{2}{*}{$\sigma=0.01$}  & GPDNet MSE & 28.27 & 24.43 & \textbf{23.69} & 23.84 \\ \cline{2-6} 
  & GPDNet MSE-SP & 30.38 & 25.54 & 24.31 & 24.44 \\ \hline
\multirow{2}{*}{$\sigma=0.015$} & GPDNet MSE & 40.46 & 35.68 & 36.09 & 36.67 \\ \cline{2-6} 
  & GPDNet MSE-SP & 46.05 & 36.99 & \textbf{35.39} & 35.80 \\ \hline
\multirow{2}{*}{$\sigma=0.02$} & GPDNet MSE & 58.88 & 50.34 & 52.96 & 55.45 \\ \cline{2-6} 
  & GPDNet MSE-SP & 64.63 & 51.82 & \textbf{49.26} & 50.43 \\ \hline
\end{tabular}
\label{table:nn}
\vspace{-1cm}
\end{table}

\begin{table}[]
\centering
\caption{Number of neighbors (UNAE - degrees).}
\setlength\tabcolsep{5pt} 
\vspace{-0.2cm}
\begin{tabular}{clcccc}
  &  & 4-NN & 8-NN & 16-NN & 24-NN \\ \hline \hline
\multirow{2}{*}{$\sigma=0.01$}  & GPDNet MSE & 27.22 & 22.51 & \textbf{20.11} & 20.89 \\ \cline{2-6} 
  & GPDNet MSE-SP & 29.04 & 24.10 & 22.33 &  22.16\\ \hline
\multirow{2}{*}{$\sigma=0.015$} & GPDNet MSE & 28.03 & 23.56 & \textbf{21.16} & 21.18 \\ 
\cline{2-6} 
  & GPDNet MSE-SP & 31.31 & 26.29 & 24.46 & 23.80 \\ \hline
\multirow{2}{*}{$\sigma=0.02$} & GPDNet MSE & 31.09 & 25.67 & \textbf{22.78} & 22.98 \\ \cline{2-6} 
  & GPDNet MSE-SP & 32.00 &  28.81& 27.06 & 26.92 \\ \hline
\end{tabular}
\label{table:nn_norm}
\vspace{-1cm}
\end{table}

\subsection{Feature analysis}

\begin{figure*}
\centering
\includegraphics[width=0.29\textwidth]{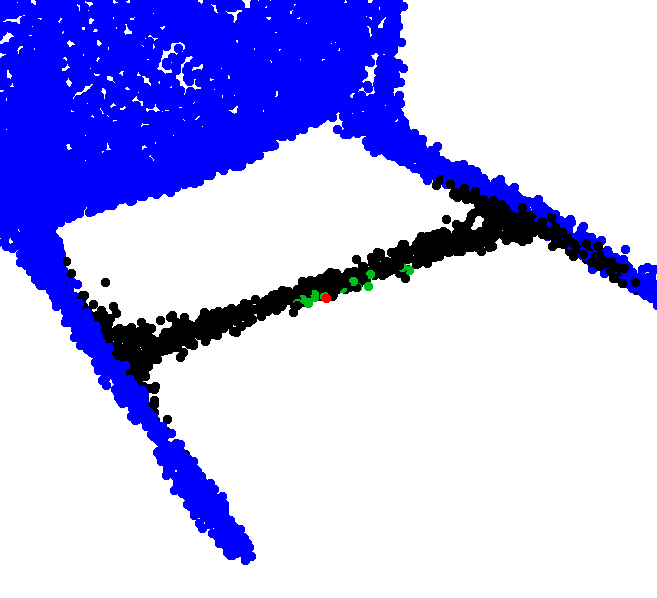}
\includegraphics[width=0.29\textwidth]{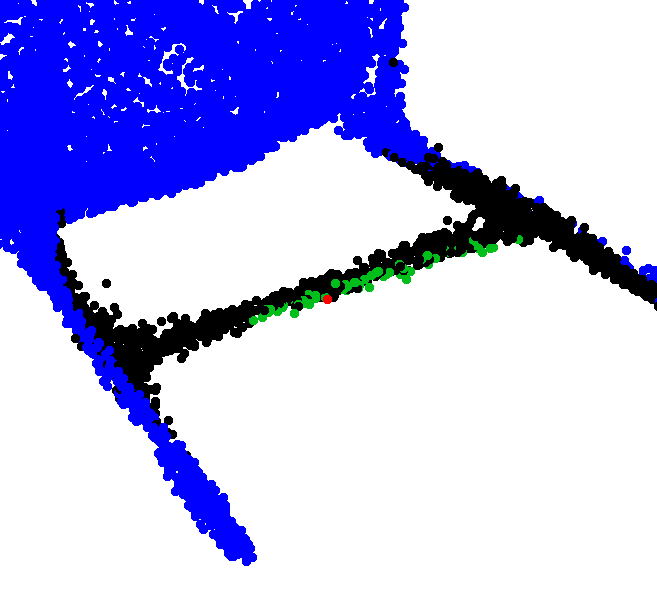}
\includegraphics[width=0.29\textwidth]{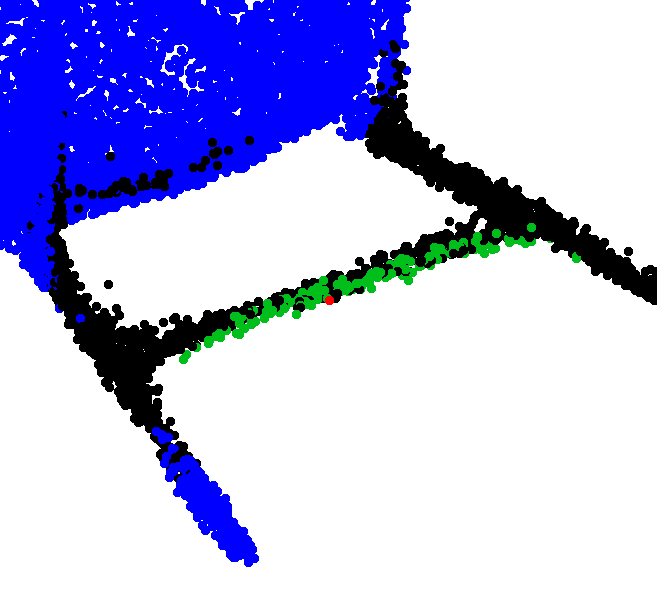}
\vspace{-0.3cm}
\caption{Receptive field (green) and search area (black) of a point (red) for the output of the three graph-convolutional layers of the second residual block of the network with respect to the input of the first graph-convolutional layer in the block. Effective receptive field size: 16, 65, 189 points.}
\label{fig:recfield}
\end{figure*}

We analyze the characteristics of the receptive field, i.e., the set of points whose feature vectors influence the features of a specific point, induced by the graph convolutional layers. In Fig. \ref{fig:recfield} we show an example of the receptive field of a single point for the output of the graph convolutional layers of a residual block with respect to the input of the residual block. The visualization is on the denoised point cloud. We observe that the receptive field is quite localized in the 3D space and its size increases as the number of layers increases. It is interesting to note that, since the graph is dynamically constructed in the feature space, the points of the receptive field are not just the spatially closest ones but they are also among the ones with similar shape characteristics. 
For example, in Fig. \ref{fig:recfield} the considered point is on the lower side of the chair stretcher and all the points of the receptive field belong to the same part of the surface.

\begin{figure*}
     \centering
     \includegraphics[width=0.49\columnwidth]{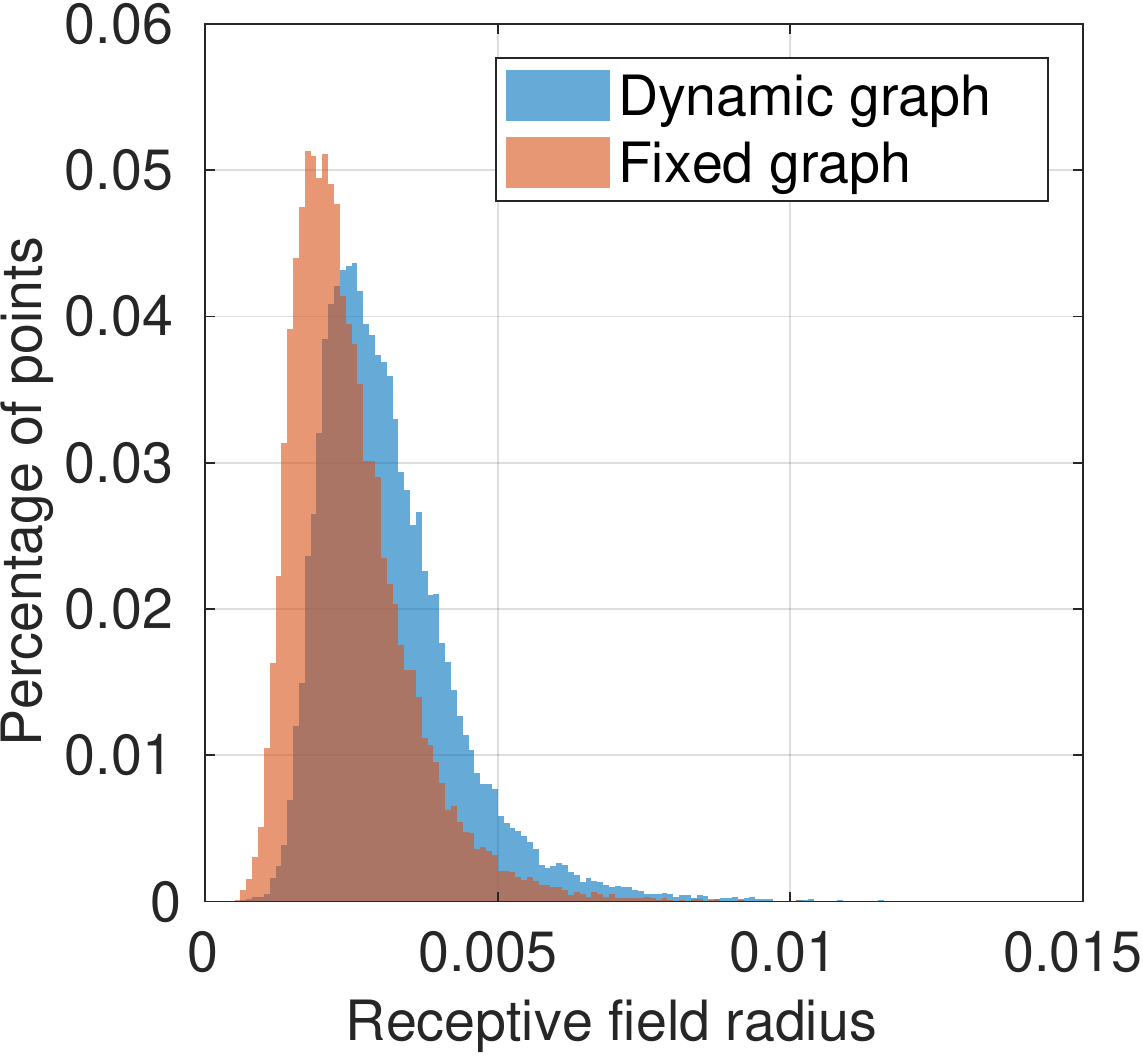}
     \includegraphics[width=0.49\columnwidth]{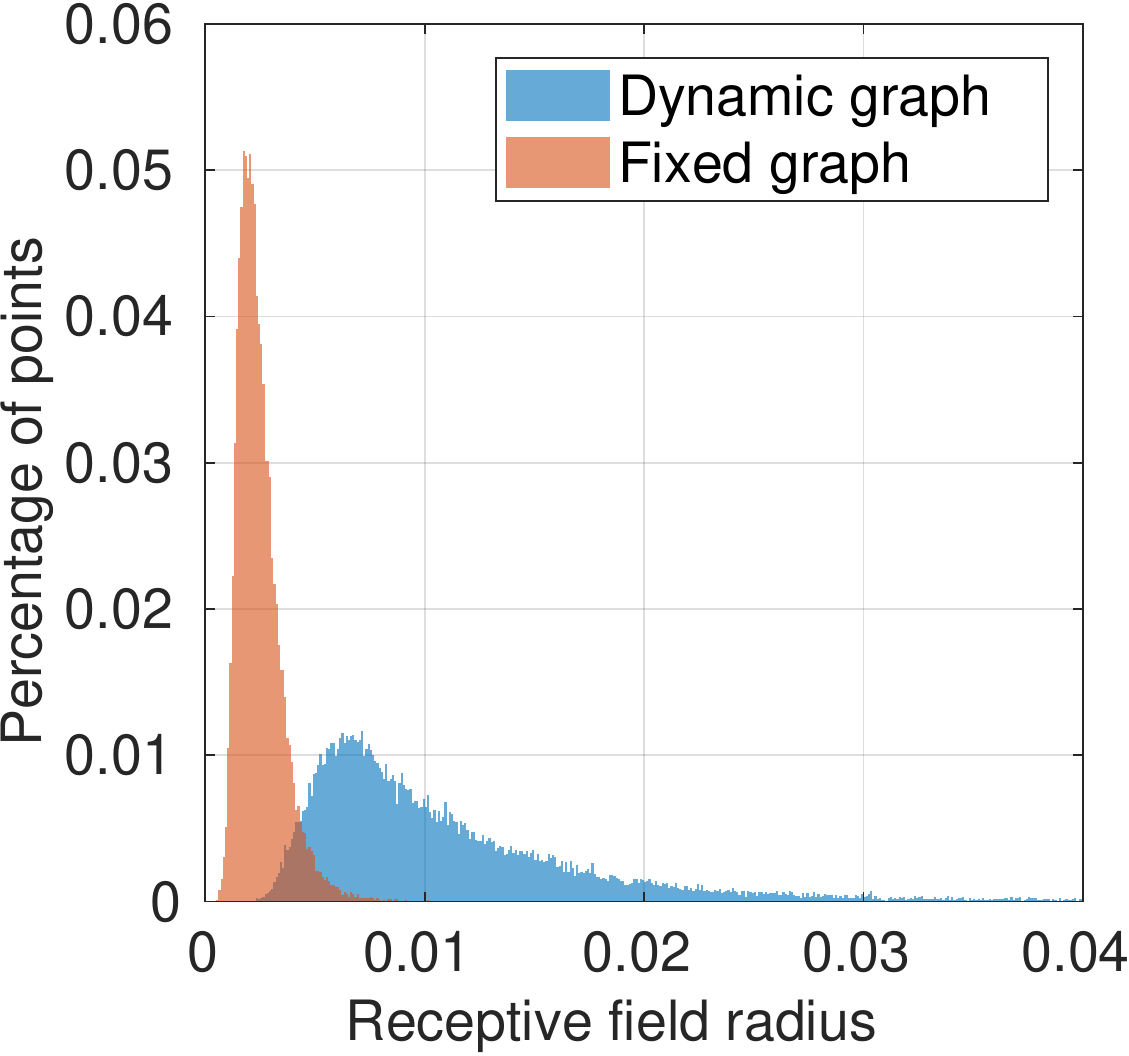}
     \caption{Radius of receptive field of points at the output of residual block with respect to its input. Left: first residual block. Right: second residual block. Neighbor selection in the noisy 3D space for fixed graph and in the feature space for dynamic graph. Radius is measured as the 90 percentile Euclidean distance to the points in the receptive field on the clean 3D point cloud.}
     \label{fig:size_receptive_dynfix}
\end{figure*}

In order to better analyze this non-local property of the receptive field we measure its radius in the 3D space and compare it to a fixed graph construction where the neighbors are determined by proximity in the noisy 3D space.
Fig. \ref{fig:size_receptive_dynfix} shows the radius of the receptive field of each point at the output of a residual block with respect to the input of the residual block. The radius is evaluated as the 90 percentile Euclidean distance in the 3D space on the clean point cloud (90 percentile is used since the maximum might be an unstable metric). It can be noticed that when using the dynamic graph construction the radius is only slightly larger in the first residual block but can be significantly larger in the second one. This can be interpreted as the feature space building and exploiting more and more non-local features with patterns similar to those in Fig. \ref{fig:recfield}.

\subsection{Structured noise}

\begin{table}[]
\centering
\caption{Velodyne scan structured noise, RMSD, 8-NN.}
\setlength\tabcolsep{3pt} 
\vspace{-0.1cm}
\begin{tabular}{cccc}
Noisy  & PointCleanNet & \textbf{GPDNet MSE} & \textbf{GPDNet MSE-SP} \\ \hline \hline
0.1447 & 0.0966 & 0.0664 & \textbf{0.0602} \\ \hline
\end{tabular}
\label{table:velodyne}
\end{table}

In order to check if the proposed architecture can generalize beyond white Gaussian noise, we train it on a simulated LiDAR dataset. We simulate scanning the Shapenet objects with a Velodyne HDL-64E scanner using the Blensor software \cite{blensor}. Two sources of noise are considered for the acquisition process: a laser distance bias with Gaussian distribution and a per-ray Gaussian noise. We set both distributions to be zero-mean and with a standard deviation equal to $1\%$ of the longest side of the object bounding box. We also retrained PointCleanNet on the simulated data for comparison with a state-of-the-art model. Table \ref{table:velodyne} shows that the results follow those on white Gaussian noise, with the proposed method improving over PointCleanNet. Note that RMSD is used as metric in place of the Chamfer measure since it is better suited to the case when points are not uniformly distributed.

\section{Conclusions}
In this paper, we have presented a new graph-convolutional neural network targeted for point cloud denoising. Thanks to the graph-convolutional layers, the proposed architecture is fully convolutional and can learn hierarchies of features, showing a behaviour similar to standard CNNs. The experimental results show that the proposed method provides a significant improvement over state-of-the-art techniques. In particular, the proposed method is robust to high level of noise and structured noise distributions, such as those observed in real LiDAR scans.

\subsubsection*{Acknowledgements.}
This material is based upon work supported by Google Cloud.

\clearpage
%
%
\bibliographystyle{splncs}
\bibliography{biblio}
\end{document}